\documentclass[manuscript]{acmart}

\AtBeginDocument{%
  \providecommand\BibTeX{{%
    \normalfont B\kern-0.5em{\scshape i\kern-0.25em b}\kern-0.8em\TeX}}}

\setcopyright{acmcopyright}
\copyrightyear{2018}
\acmYear{2018}
\acmDOI{XXXXXXX.XXXXXXX}

\usepackage{multirow}
\usepackage{colortbl}
\usepackage{color}
\usepackage{graphicx}

\copyrightyear{2023}
\acmYear{2023}
\setcopyright{acmlicensed}\acmConference[UMAP '23 Adjunct]{Adjunct Proceedings of the 31st ACM Conference on User Modeling, Adaptation and Personalization}{June 26--29, 2023}{Limassol, Cyprus}
\acmBooktitle{Adjunct Proceedings of the 31st ACM Conference on User Modeling, Adaptation and Personalization (UMAP '23 Adjunct), June 26--29, 2023, Limassol, Cyprus}
\acmPrice{15.00}
\acmDOI{10.1145/3563359.3597390}
\acmISBN{978-1-4503-9891-6/23/06}

\begin{document}

\title[Less Can Be More]{Less Can Be More: Exploring Population Rating Dispositions with Partitioned Models in Recommender Systems}

\author{Ruixuan Sun}
\affiliation{%
 \institution{Grouplens Research, University of Minnesota}
 \streetaddress{5-244 Keller Hall, 200 Union Street SE}
 \city{Minneapolis}
 \state{Minnesota}
 \country{United States}}

\author{Ruoyan Kong}
\affiliation{%
 \institution{Grouplens Research, University of Minnesota}
 \streetaddress{5-244 Keller Hall, 200 Union Street SE}
 \city{Minneapolis}
 \state{Minnesota}
 \country{United States}}

\author{Qiao Jin}
\affiliation{%
 \institution{Grouplens Research, University of Minnesota}
 \streetaddress{5-244 Keller Hall, 200 Union Street SE}
 \city{Minneapolis}
 \state{Minnesota}
 \country{United States}}

\author{Joseph A. Konstan}
\affiliation{%
 \institution{Grouplens Research, University of Minnesota}
 \streetaddress{5-244 Keller Hall, 200 Union Street SE}
 \city{Minneapolis}
 \state{Minnesota}
 \country{United States}}

\renewcommand{\shortauthors}{Sun, et al.}

\begin{abstract}

 In this study, we partition users by rating disposition - looking first at their percentage of negative ratings, and then at the general use of the rating scale. We hypothesize that users with different rating dispositions may use the recommender system differently and therefore the agreement with their past ratings may be less predictive of the future agreement. 
 
 We use data from a large movie rating website to explore whether users should be grouped by disposition, focusing on identifying their various rating distributions that may hurt recommender effectiveness. We find that such partitioning not only improves computational efficiency but also improves top-k performance and predictive accuracy. Though such effects are largest for the user-based KNN CF, smaller for item-based KNN CF, and smallest for latent factor algorithms such as SVD.
 
\end{abstract}



\maketitle

\section{Introduction}
  
  Collaborative filtering recommenders are based on the idea that each user benefits from recommendations based on the ratings of other users. Implicitly, this means that the other users' data are in some way appropriate to use -- that users base their ratings on some common value system (even if they have different preferences) or that past agreement is predictive of the future agreement. We already know that this assumption doesn't extend arbitrarily. Prior work has shown that in those cases it may result in better recommendations if the rating dataset is partitioned based on topic modeling and users' persona \cite{wilson2014improving}. 
  
  Currently, we know 3 different biases that control what a rating looks like - 1) Selection bias arises from user choices of items to consume. Typically, users prefer items that they anticipate enjoying \cite{dalvi_para_2013}. 2) Under-reporting bias results from users' decisions of which items to rate.  Empirical studies suggest that most users' average ratings are close to or even higher than the scale's average, as consumers are more likely to provide feedback on items they feel positively about. Additionally, users are more likely to rate items if they are either satisfied or dissatisfied rather than if they feel neutral about the product \cite{hu_self-selection_2017}. 3) Rating scale bias reflects personal differences in the use of rating scales across users \cite{gena_impact_2011,cena_how_2017}.
  
  Looking at data from a movie recommender, we wonder whether it might also be useful to partition users who come to the recommender with very different goals and ways of using the system. We call this way of using the recommender system -- the choice of items to rate and how they're rated -- the user's rating disposition, which encapsulates the three biases mentioned above. We want to explore the usage patterns by implicitly clustering users based on empirical distributions of their ratings. We frame our study into three research questions:
 
 \noindent\textbf{RQ1}: How to categorize different user populations based on their rating distribution behaviors?
 
 \noindent\textbf{RQ2}: What are the impacts of ratings from different user populations on other populations?
 
 \noindent\textbf{RQ3}: Can we improve the recommendation performance by using fewer data from different populations based on their individual partitioned models?
 
 To answer RQ1, we first split users into two populations based on their negative rating percentages. We also further categorize more granular subgroups based on positive rating peak distribution for two populations. Built upon the findings from RQ1, we run a preliminary partitioning experiment on two different populations for RQ2 by training the recommender model separately for optimistic and pessimistic users. Finally, we move to RQ3 with more empirical trials and show that we can potentially improve the recommendation quality and better serve each population with partitioned model training. 
 
 In the following sections, we first discuss related work, then we introduce our data collection and rating disposition identification process. A preliminary binary population partitioning trial and more thorough partitioned modeling are then demonstrated, and we conclude with discussion and potential future implications.
 
\section{Related Work}
\subsection{Rating Scale Inconsistency}
 Each user has a personal rating scale. Some people like to give 5 out of 5 stars to express their preferences on items they enjoy, while others might be more conservative and just give 3 out of 5 stars. Many mathematical models use normalization or z-score techniques \cite{herlocker2002empirical} to capture each user's average rating information to mitigate their rating scale discrepancy. \citeauthor{gena_impact_2011} and \citeauthor{cena_how_2017} demonstrated that rating scales have their own “personality” and contain user behavior information that also should be incorporated into the recommendation generation process \cite{gena_impact_2011,cena_how_2017}. \citeauthor{nguyen2013rating} have studied how to overcome the rating inconsistency for individual users by introducing interface examples during rating process\cite{nguyen2013rating}. \citeauthor{said2012users} estimated a magic barrier with noise from user rating inconsistency to assess the actual quality of recommender systems \cite{said2012users}. In this study, we seek to group users with similar scale standards together to avoid personal scale bias.

\subsection{Selection Bias}
 In most cases, negative ratings just constitute a small proportion in recommender system databases. \citeauthor{dalvi_para_2013} proposed the idea of selection bias in the numerical rating scenarios, showing that average user rating is usually of a positive and skewed distribution\cite{dalvi_para_2013}. \citeauthor{hu2009overcoming} mentioned the J-shaped distribution of product reviews mainly contain most positive and small amount of extremely negative ratings\cite{hu2009overcoming}.
 Studies from \citeauthor{meijerink_why_2020} strengthened the distribution concern with Airbnb data analysis\cite{meijerink_why_2020} by pointing out the under-reporting on scale. They indicated that users would be more likely to leave a review if the service quality they received is good, therefore making the ratings and reviews skewed positive. Many domains of recommendation are overflowed with positive ratings. However, rare events like negative ratings actually contain more information content \cite{arndt_information_2003,kong2021learning}. They are interesting to study not only for the content itself but also for the habits of using the system behind the user ratings. As \citeauthor{zeng_negative_2011} mentioned, negative ratings from less active users to less popular objects could probably have a positive impact on the recommendations\cite{zeng_negative_2011}. In Section \ref{section4}, we describe methods to identify negative users based on their rating behaviors and evaluate the potential bias from different user groups to another.
 
 \subsection{Gaps}
 
 Though many previous studies have implemented many data-driven approaches by collecting user engagement \cite{he2023hiercat,kong2022multi} or real-time interaction \cite{kong2023getting} data to understand the effectiveness of recommender systems, we propose to extract patterns of rating disposition from just the most commonly available rating data, which is usually available for researchers from open-sourced datasets. Inferring from users’ consumption habits based on features of their rating distribution, we map user consumption and rating patterns together and build partitioned models that can improve the performance of recommender systems.

\section{Data Collection}
  We use data from the movie recommender system MovieLens (https://movielens.org). Movielens is a non-commercial, personalized recommender that gathers users’ ratings on movies they have watched and provides predictions of new movies with different recommendation algorithms. Our data contains 3,908,657 ratings from 6724 users on 68,044 movies. Those are active users who have logged in to the website over 12 times from 2019 to 2020 and rated over 20 movies since their registration. Administrative users are excluded. The dataset can be downloaded at https://grouplens.org/datasets/rating-disposition-2023.

\section{Identify Rating Disposition}\label{section4}

 Our first approach for quantifying rating disposition is to categorize users based on their negative rating percentage. Previous studies usually treated 1-star or 2-star as negative ratings on a 5-star scale \cite{zeng_negative_2011}. In our dataset, the median of rating is 3.5, so we adjust the definition based on our average and define the negative rating as less than or equal to 3 stars, and positive ratings as >=3.5 stars, since Movielens only has half-star intervals. After that, we introduce the concept of optimistic population and pessimistic population.
 
\begin{figure}[!htbp]
    \centering
      \includegraphics[width=0.3\columnwidth]{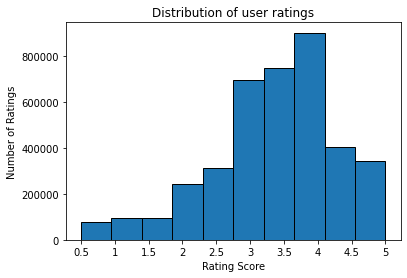}
      \includegraphics[width=0.3\columnwidth]{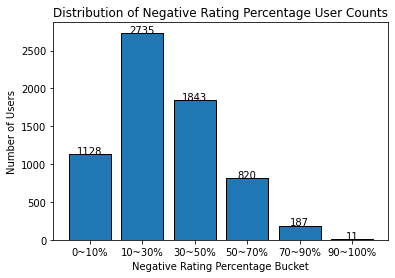}
      \includegraphics[width=0.3\columnwidth]{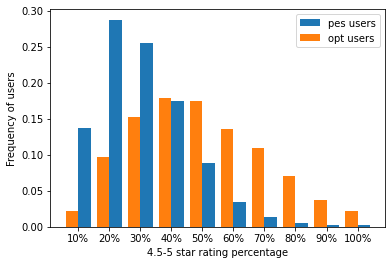}
      \caption{The distribution of user ratings, negative rating percentage buckets, and extremely positive ratings for two groups of users}~\label{fig:population_rating_characteristics}
    \end{figure}

 To find out the threshold for splitting two different populations, we compute the percentage of negative ratings among all ratings for each user, and then aggregate individual data into six different groups: [0-10\%, 10-30\%, 30-50\%, 50-70\%, 70-90\%, 90-100\%], each indicating the percentage bucket of negative ratings. The justification for the non-equal bin width is to segregate out the most optimistic (0-10\%) and pessimistic (90-100\%) user groups and potentially test their impact on other groups, even if their absolute number of users is small. Based on the distribution observation, 57.5\% users have less than or equal to 30\% ratings among all of their ratings less than or equal to 3 stars, while 42.5\% users have more than 30\% negative ratings. For better representation and split equality, we choose 30\% as the optimistic and pessimistic user threshold and hypothesize that users with different percentage of negative ratings reflect their potentially dissimilar dispositions. Based on the hypothesis, we also plot out the ratio of extremely positive ratings (4.5-star and 5-star) over all positive ratings (>= 3.5 stars) for both populations in the last graph of Fig. \ref{fig:population_rating_characteristics}, which shows that pessimistic users give fewer high ratings and potentially have different positive rating scales from optimistic users. 
 
 In the next section, we show how to use the positive rating peak along with the optimistic and pessimistic population definitions to categorize more detailed disposition types.

\section{Potential Improvements by Model Partitioning}
 In this section, we seek to mimic a more efficient recommendation model that only trains and tests each different user population within their own group of neighbors. We hypothesize that the data from each disposition group will be most relevant to recommendations to the save group and possibly harmful to other population group recommendations.
 
 We conduct three different experiments to evaluate recommendation model performance on the original data as well as on the partitioned data based on our definition of populations. We choose three classic recommender algorithms to evaluate model performance: User-User CF, Item-Item CF, and the latent factored CF using matrix factorization approximation Funk SVD. All algorithms are impelmented with the Surprise API\cite{hug2020surprise}. For User-user and Item-item CF, they both make use of \emph{KNNWithMeans} model, with parameters \emph{min\_K} = 1, \emph{max\_K} = 20, and similarity measure = Mean Squared Difference (MSD). For the SVD algorithm, we set \emph{n\_factors} = 10, and \emph{n\_epochs} = 20. We note that different algorithms may be more or less robust to having users from different rating dispositions together. We hypothesize that User-User CF will be highly sensitive to impacts from their nearest neighbors and more likely to benefit from partitioning, since users with different dispositions are rarely neighbors. But it also could be that they often become "false neighbors" that correlate well on the data they have rated in common but then turn out to disagree badly on other items. By contrast, latent factor methods such as SVD create a single dimensionalization (the latent factors) from all users with consideration of model user-specific bias -- having different subpopulations might have little effect because they can all be served by the same factors even if they weigh different factors differently, or having them together might result in a dimensionalization that's worse for each group. We carry out experiments to explore these questions.
 
\subsection{Optimistic and Pessimistic Group Partition}

\begin{figure}[!htbp]
    \centering
      \includegraphics[width=0.8\columnwidth]{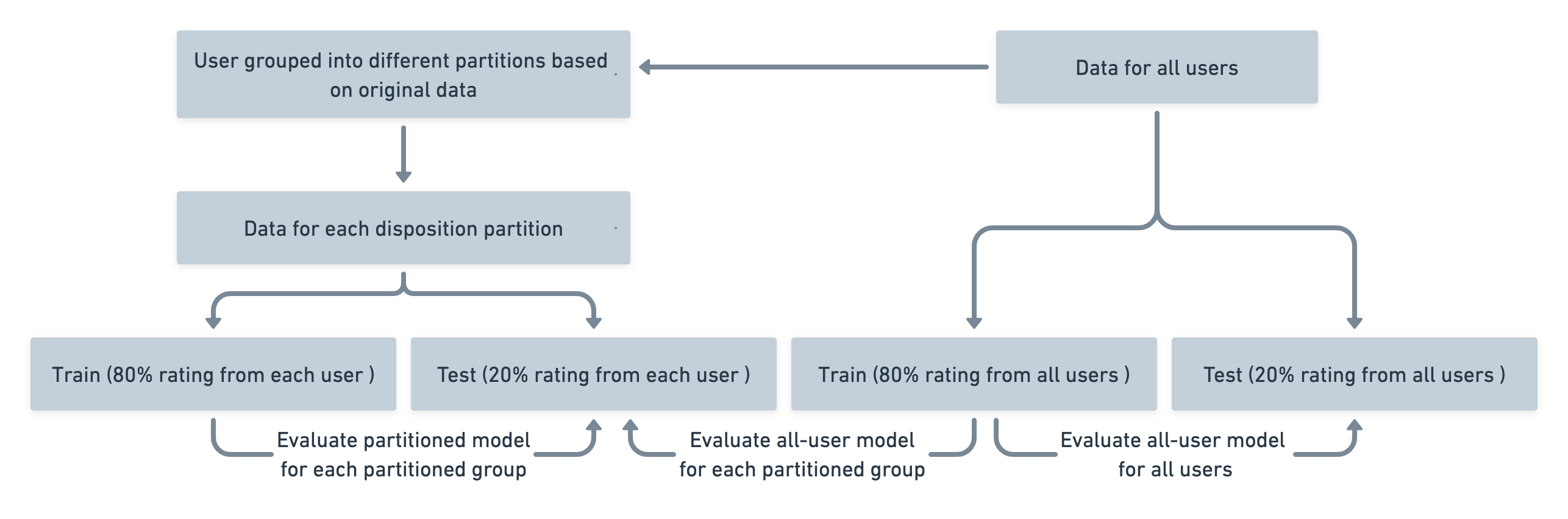}
      \caption{The general model construction steps}~\label{fig:model_construction_flow}
    \end{figure}

  We first extract optimistic and pessimistic populations based on each user's whole rating profile and train the two partitioned models on individual test sets to assess their performances. For each user group, we conduct a 5-fold cross-validation and report the average result of metrics. The detailed model construction steps are shown in Fig. \ref{fig:model_construction_flow}.
  
  The evaluation metrics we utilize are normalized discounted cumulative gain(NDCG@8), precision@8, and RMSE. We evaluate both NDCG and precision based on the top 8 recommended items since it is the standard size of displayed recommendation list on the Movielens website. For NDCG@8, we add an equal number of unrated movies to the raw test set to model the real-world recommendation scenario, since in practice users might find the recommended movies not necessarily the ones they have watched or rated. In general, higher NDCG and precision metrics indicate better top-k item recommendation relevance and true-positive rate, while a lower RMSE value means we have a lower error rate. For the significance test, we compare the 5-fold metrics result from each partitioned model to the original model, and adjust the p-value for each of the three metrics with Bonferroni correction to reduce the false positive rate. \cite{weisstein2004bonferroni}

\begin{table}
\centering
\resizebox{\linewidth}{!}{%
\begin{tabular}{cccccccccc} 
\hline
\multirow{2}{*}{Train\textbackslash{}Test} & \multicolumn{3}{c}{all users (6724 users)} & \multicolumn{3}{c}{optimistic (3863 users)}   & \multicolumn{3}{c}{pessimistic (2861 users)}                              \\
                                           & ndcg@8     & precision@8 & rmse            & ndcg@8         & precision@8 & rmse           & ndcg@8                 & precision@8            & rmse                    \\
all users                                  & 0.479      & 0.483       & 0.854           & 0.643          & 0.669       & 0.756          & 0.288                  & 0.274                  & 0.909                   \\
optimistic                                 & \textbf{-} & -           & -               & \textbf{0.644} & 0.668       & \textbf{0.752} & -                      & -                      & -                       \\
pessimistic                                & \textbf{-} & -           & -               & -              & -           & -              & \textbf{0.332*(0.022)} & \textbf{0.323*(0.000)} & \textbf{0.901*(0.003)}  \\
\hline
\end{tabular}
}
\caption{The performance of the partitioned model for optimistic and pessimistic users on its own test set with User-User CF. The improved metric is indicated with bold font, and the significant paired t-test is indicated by *(pval). }
\label{tab:population_partition}
\end{table}

 Table \ref{tab:population_partition} displays that using User-User CF to build separate recommendation models for both optimistic and pessimistic users improves NDCG@8 and RMSE performances, while the precision value for pessimistic partitioning outperforms that from the all-user model with statistical significance. This fact implies that training with all raw data potentially introduce harmful bias to different population groups, and removing the less relevant part can effectively reduce the disposition bias,\footnote{We have not reported the Item-Item CF and SVD results as they do not show significant impacts on partitioned models.} which helps support our hypothesis of the existence of rating disposition in recommender systems.
 
\subsection{Negative Percentage Group Partition}

The second empirical study is an expansion of the first one. We further categorize different user populations based on their negative rating percentage and test their partitioning performances on their own test set. With the original "all users" model performance in Table \ref{tab:neg_percentage_partition}, we witness the consequence of selection bias - those with more percentage of negative ratings have worse performance compared to the ones who like to give a higher ratio of positive ratings.
 
 On different negative percentage populations, we can see that the partitioning works best in User-User CF, where almost all groups witness significantly improved NDCG and precision scores compared to the baseline original model. The RMSE score dropped for 4 out of 6 percentage groups but in a very trivial way, indicating we are not compromising the accuracy of recommendations with partitioned models even with far less training data compared to the original training set. We also notice that the more pessimistic groups with negative rating percentages larger than 30\% see more significant metrics improvement compared to the optimistic groups. In Item-Item CF, the two pessimistic groups 70-90\% and 90-100\% benefit more from training on their own data, while the other percentage groups witnessed a slight drop in recommendation quality. As for the SVD algorithm, some positive impact is detected on the optimistic  0-10\% group, while others do not seem to be improved from partitioning.

\begin{table}
\centering
\resizebox{\linewidth}{!}{%
\begin{tabular}{cccccccc} 
\hline
\multicolumn{2}{c}{Algorithm}                         & \multicolumn{2}{c}{User-User CF}   & \multicolumn{2}{c}{Item-Item CF} & \multicolumn{2}{c}{SVD}     \\
\multicolumn{2}{c}{Test\textbackslash{}Train}         & all users & partitioned            & all users & partitioned          & all users & partitioned     \\ 
\hline
\multirow{3}{*}{all users (6724 users)} & ndcg@8      & 0.456     & \textbf{-}             & 0.515     & -                    & 0.758     & -               \\
                                        & precision@8 & 0.449     & -                      & 0.519     & -                    & 0.755     & -               \\
                                        & rmse        & 0.854     & -                      & 0.876     & -                    & 0.799     & -               \\ 
\hline
\multirow{3}{*}{0-10\% (1128 users)}    & ndcg@8      & 0.766     & \textbf{0.778}         & 0.792     & 0.776                & 0.883     & \textbf{0.887}  \\
                                        & precision@8 & 0.800     & \textbf{0.806}         & 0.824     & 0.805                & 0.905     & \textbf{0.913}  \\
                                        & rmse        & 0.632     & \textbf{0.622}         & 0.642     & 0.633                & 0.582     & 0.607           \\ 
\hline
\multirow{3}{*}{10-30\% (2735 users)}   & ndcg@8      & 0.609     & \textbf{0.633}         & 0.657     & 0.636                & 0.843     & 0.824           \\
                                        & precision@8 & 0.622     & \textbf{0.650}         & 0.682     & 0.653                & 0.864     & 0.846           \\
                                        & rmse        & 0.770     & 0.776                  & 0.782     & 0.790                & 0.723     & 0.752           \\ 
\hline
\multirow{3}{*}{30-50\% (1843 users)}   & ndcg@8      & 0.399     & \textbf{0.458}         & 0.482     & 0.470                & 0.783     & 0.763           \\
                                        & precision@8 & 0.388     & \textbf{0.460}         & 0.488     & 0.463                & 0.781     & 0.757           \\
                                        & rmse        & 0.875     & 0.880                  & 0.895     & 0.906                & 0.825     & 0.852           \\ 
\hline
\multirow{3}{*}{50-70\% (820 users)}    & ndcg@8      & 0.159     & \textbf{0.246*(0.030)} & 0.279     & 0.275                & 0.693     & 0.677           \\
                                        & precision@8 & 0.143     & \textbf{0.227*(0.007)} & 0.272     & 0.238                & 0.650     & 0.620           \\
                                        & rmse        & 0.926     & 0.934*(0.006)          & 0.958     & 0.970                & 0.865     & 0.909           \\ 
\hline
\multirow{3}{*}{70-90\% (187 users)}    & ndcg@8      & 0.093     & \textbf{0.246}         & 0.201     & \textbf{0.278}       & 0.624     & 0.621           \\
                                        & precision@8 & 0.073     & \textbf{0.145}         & 0.155     & \textbf{0.164}       & 0.493     & 0.427           \\
                                        & rmse        & 1.009     & 1.041                  & 1.047     & 1.084                & 0.902     & 0.978           \\ 
\hline
\multirow{3}{*}{90-100\% (11 users)}    & ndcg@8      & 0.087     & \textbf{0.307}         & 0.120     & \textbf{0.313}       & 0.422     & 0.378           \\
                                        & precision@8 & 0.036     & \textbf{0.066}         & 0.045     & \textbf{0.066}       & 0.193     & 0.084           \\
                                        & rmse        & 1.055     & \textbf{1.014}         & 1.114     & \textbf{1.009}       & 0.909     & 0.944           \\
\hline
\end{tabular}
}
\caption{The performance of the partitioned model for negative percentage users on its own test set. In the column index, “partitioned” means each negative percentage training set matching the row name. The improved metric is indicated with bold font, and the significant paired t-test is indicated by *(pval). }
\label{tab:neg_percentage_partition}
\end{table}

\subsection{Positive Rating Peak Group Partition}

  Another partitioned model is built with the positive rating peak characteristics. This experiment split users into 6 groups, including all users, optimistic users (their positive rating peak is at 4 stars), and pessimistic users with their peak positive ratings at each half-star within the range of [3.5, 5]. As shown in Table \ref{tab:pos_rating_partition}, it proves to be working best with user-user CF, where all populations witnessed an increase in NDCG and precision metrics with statistic significance. The way different populations tend to give high positive ratings is quite different and can be a potential contributor to bias when their ratings are included with other groups. We also observe slight improvement for item-item CF when the peak is at 3.5 stars and 4.5 stars as well, but both groups also suffer a certain level of drop in error rate. Like the previous group of experiments, the SVD model does not seem to benefit from the rating peak group partitioning.

\begin{table}
\centering
\resizebox{\linewidth}{!}{%
\begin{tabular}{cccccccc} 
\hline
\multicolumn{2}{c}{Algorithm}                                                                                     & \multicolumn{2}{c}{User-User CF}   & \multicolumn{2}{c}{Item-Item CF}   & \multicolumn{2}{c}{SVD}  \\
\multicolumn{2}{c}{Test\textbackslash{}Train}                                                                     & all users & partitioned            & all users & partitioned            & all users & partitioned  \\ 
\hline
\multirow{3}{*}{\begin{tabular}[c]{@{}c@{}}all users\\(6724 users)\end{tabular}}                    & ndcg@8      & 0.214     & -                      & 0.418     & -                      & 0.832     & -            \\
                                                                                                    & precision@8 & 0.220     & -                      & 0.418     & -                      & 0.741     & -            \\
                                                                                                    & rmse        & 0.812     & -                      & 0.823     & -                      & 0.753     & -            \\ 
\hline
\multirow{3}{*}{\begin{tabular}[c]{@{}c@{}}optimistic\\(3862 users)\end{tabular}}                   & ndcg@8      & 0.273     & \textbf{0.358*(0.000)} & 0.487     & 0.419                  & 0.854     & 0.832        \\
                                                                                                    & precision@8 & 0.294     & \textbf{0.378*(0.000)} & 0.509     & 0.441                  & 0.804     & 0.779        \\
                                                                                                    & rmse        & 0.708     & 0.715*(0.000)          & 0.703     & 0.725                  & 0.663     & 0.689        \\ 
\hline
\multirow{3}{*}{\begin{tabular}[c]{@{}c@{}}pessimistic with \\peak@3.5\\ (1403 users)\end{tabular}} & ndcg@8      & 0.122     & \textbf{0.272*(0.000)} & 0.285     & \textbf{0.317*(0.001)} & 0.797     & 0.796        \\
                                                                                                    & precision@8 & 0.098     & \textbf{0.231*(0.000)} & 0.239     & \textbf{0.247*(0.018)} & 0.602     & 0.590        \\
                                                                                                    & rmse        & 0.779     & \textbf{0.771*(0.000)} & 0.799     & 0.802*(0.000)          & 0.718     & 0.737        \\ 
\hline
\multirow{3}{*}{\begin{tabular}[c]{@{}c@{}}pessimistic with\\peak@4.0\\ (1257 users)\end{tabular}}  & ndcg@8      & 0.137     & \textbf{0.276*(0.000)} & 0.342     & 0.307                  & 0.807     & 0.806        \\
                                                                                                    & precision@8 & 0.132     & \textbf{0.268*(0.000)} & 0.332     & 0.283                  & 0.711     & 0.698        \\
                                                                                                    & rmse        & 0.914     & 0.918*(0.005)          & 0.925     & 0.954                  & 0.847     & 0.877        \\ 
\hline
\multirow{3}{*}{\begin{tabular}[c]{@{}c@{}}pessimistic with\\peak@4.5\\ (32 users)\end{tabular}}    & ndcg@8      & 0.278     & \textbf{0.665*(0.000)} & 0.537     & \textbf{0.655}         & 0.826     & 0.757        \\
                                                                                                    & precision@8 & 0.251     & \textbf{0.501*(0.001)} & 0.476     & \textbf{0.484}         & 0.686     & 0.575        \\
                                                                                                    & rmse        & 1.150     & 1.334*(0.000)          & 1.149     & 1.307                  & 1.088     & 1.226        \\ 
\hline
\multirow{3}{*}{\begin{tabular}[c]{@{}c@{}}pessimistic with\\peak@5.0\\ (160 users)\end{tabular}}   & ndcg@8      & 0.209     & \textbf{0.434*(0.000)} & 0.484     & 0.465                  & 0.806     & 0.768        \\
                                                                                                    & precision@8 & 0.187     & \textbf{0.388*(0.000)} & 0.437     & 0.404                  & 0.686     & 0.612        \\
                                                                                                    & rmse        & 1.373     & 1.499*(0.000)          & 1.432     & 1.536                  & 1.280     & 1.415        \\
\hline
\end{tabular}
}
\caption{The performance of the partitioned model for optimistic users and pessimistic users with different positive rating peaks on their own test set. In the column index, “partitioned” means each population training set matching the row name. The improved metric is indicated with bold font, and the significant paired t-test is indicated by *(pval). }
\label{tab:pos_rating_partition}
\end{table}

\section{Discussion}
 
  With the definition of two groups of population and other disposition indices, we identified rating disposition as a useful concept, and demonstrate its value in an online movie recommender community. We ran multiple sets of partitioning experiments limiting a model to users with the same rating disposition and explored how can we improve the recommendation models’ performance, at least for certain dispositions. The preliminary binary population model showed the benefit of building a recommender for each user group with different dispositions. We then tested with a more granular negative percentage group partitioning that had demonstrated its efficiency and strong performance boost, mainly for the User-User CF algorithms. Finally, the positive rating peak partition brought quite positive changes for the user-user CF model and also worked for certain groups in the item-item CF model, with corrected statistical significance. Compared to the results in some other previous works on de-biasing user ratings, such as the post-hoc rating adjustment technique by \citeauthor{adomavicius2014biasing}, in which the predictive accuracy (measure with RMSE) did not see significant improvement with de-biased adjustment in both user-user and item-item CF \cite{adomavicius2014biasing}, our results showed the effectiveness of population partitioning in group recommendation not only for more efficient model training with its less amount of data, but also in improving the recommendation performance with collaborative filtering algorithms. 

\section{Conclusion and Future Implications}
 In this study, we explored the patterns behind raw ratings from different user populations with a movie recommendation dataset. Our 3-step method involved identifying different rating dispositions, testing partitioned recommenders, and defining training subgroups. We demonstrate that the partitioned models can work more efficiently while also achieve positive impacts on User-User CF algorithms, and latent factor methods seem to be robust with rating dispositions. 
 
 This work is not meant to be an exhaustive exploration of group partitioning recommendation, but a case example to demonstrate the potential benefits and feasibility of the approach. We admit several limitations: 1) We focused on a single domain and limited to 5-point Likert scale; 2) We did not explore other similarity measures or more advanced algorithms; 3) We have not conducted interviews to understand users’ subjective rating preferences; 4) Our taxonomy of two user groups is simple and needs further refinement. Future work should confirm and expand our approach for a deeper understanding of when ratings disposition is useful and how to effectively segregate or exclude ratings.

\bibliographystyle{ACM-Reference-Format}
\bibliography{sample-base}

\end{document}